%% file: acllatex.tex
\pdfoutput=1

\documentclass[11pt]{article}

\usepackage[final]{acl}

\usepackage{times}
\usepackage{latexsym}

\usepackage[T1]{fontenc}

\usepackage[utf8]{inputenc}

\usepackage{microtype}

\usepackage{inconsolata}

\usepackage{graphicx}

\usepackage{url}
\usepackage{multirow}
\usepackage{multicol}
\usepackage{subcaption}
\usepackage{hyperref}
\usepackage{float}
\usepackage{amsmath}

\usepackage{lipsum}

\newcommand\blfootnote[1]{%
  \begingroup
  \renewcommand\thefootnote{}\footnote{#1}%
  \addtocounter{footnote}{-1}%
  \endgroup
}

\title{{Reverse Probing: }
{Evaluating Knowledge Transfer via Finetuned Task Embeddings for Coreference Resolution}}

\author{Tatiana Anikina$^{1,2,*}$, Arne Binder$^{1,*}$, David Harbecke$^1$, Stalin Varanasi$^{1,2}$, \\
    \textbf{Leonhard Hennig$^1$, Simon Ostermann$^{1,2}$, Sebastian Möller$^{1,3}$,} and \textbf{Josef van Genabith$^{1,2}$} \\
  $^1$German Research Centre for Artificial Intelligence, Saarbrücken \\
  $^2$Saarland Informatics Campus, Saarbrücken \\
  $^3$Technische Universität Berlin, Berlin \\
  \texttt{tatiana.anikina@dfki.de} 
}

\begin{document}
\maketitle
\begin{abstract}
In this work, we reimagine classical probing to evaluate knowledge transfer from simple source to more complex target tasks. Instead of probing frozen representations from a complex source task on diverse simple target probing tasks (as usually done in probing), we explore the effectiveness of embeddings from multiple simple source tasks on a single target task. We select coreference resolution, a linguistically complex problem requiring contextual understanding, as focus target task, and test the usefulness of embeddings from comparably simpler tasks such as paraphrase detection, named entity recognition, and relation extraction. Through systematic experiments, we evaluate the impact of individual and combined task embeddings. 

Our findings reveal that task embeddings vary significantly in utility for coreference resolution, with semantic similarity tasks (e.g., paraphrase detection) proving most beneficial. Additionally, representations from intermediate layers of fine-tuned models often outperform those from final layers. Combining embeddings from multiple tasks consistently improves performance, with attention-based aggregation yielding substantial gains. These insights shed light on relationships between task-specific representations and their adaptability to complex downstream tasks, encouraging further exploration of embedding-level task transfer. Our source code is publicly available.\footnote{\href{https://github.com/Cora4NLP/multi-task-knowledge-transfer}{github.com/Cora4NLP/multi-task-knowledge-transfer}\label{note:source-code}} \blfootnote{* Equal contribution.}
\end{abstract}

\begin{figure}[t]
    \centering
    \includegraphics[scale=0.55]{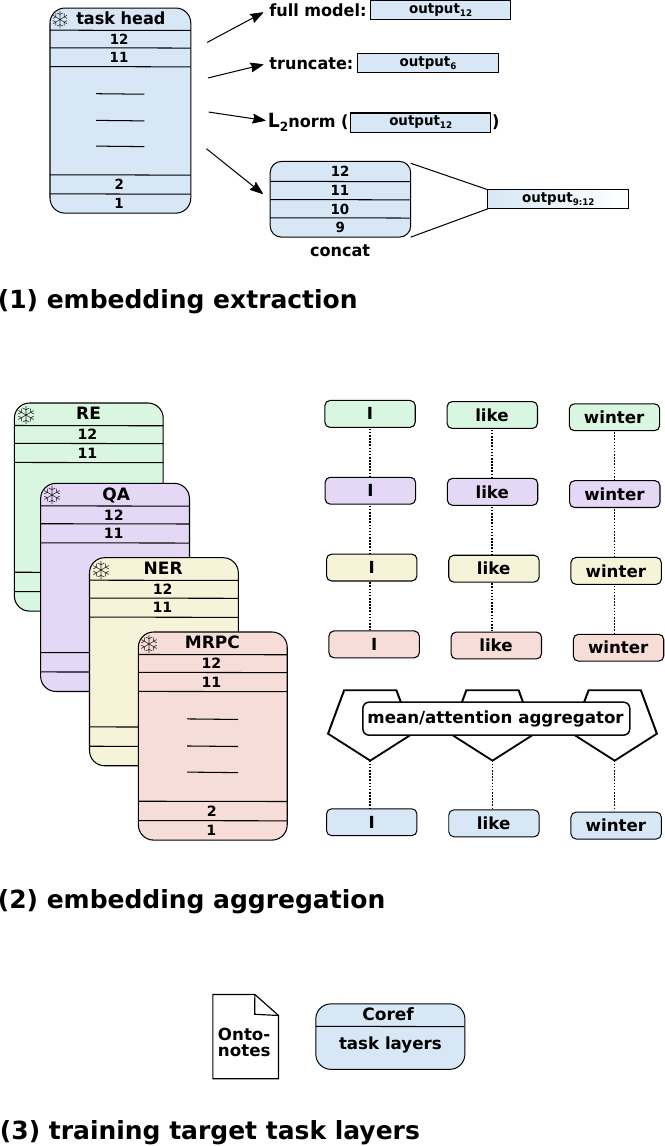}
    \caption{Probing workflow with Coreference Resolution (Coref) as target task and four different source tasks: Relation Extraction (RE), Question Answering (QA), Named Entity Recognition (NER), and Paraphrase Detection (MRPC).}
    \label{fig:probing-workflow}
    \vspace{-0.5cm}
\end{figure}

\section{Introduction}\label{sec:intro}

Language models have exhibited superior performance in most areas of NLP applications, including natural language inference \cite{N18-1101}, question answering \cite{rajpurkar-etal-2016-squad,rajpurkar-etal-2018-know}, commonsense reasoning \cite{talmor-etal-2019-commonsenseqa,ostermann-etal-2019-commonsense}, and others. Since the establishment of language models with partial super-human performance, research has aimed to pinpoint which types of knowledge are exactly encoded by such language models. One technique in the field of explainable artificial intelligence for evaluating the presence of such types of knowledge is \textit{probing} \cite{conneau-etal-2018-cram,hewitt-liang-2019-designing,tenney-etal-2019-bert,belinkov-2022-probing}. Probing involves adding linear classifiers on top of representations extracted from a pre-trained model, which are trained on simple tasks for predicting a feature of choice, such as syntactic structures \cite{lin-etal-2019-open}, entity types \cite{7d48ec9ae06d402c894e3bea30841e33}, or specific types of commonsense knowledge \cite{zhou2020evaluating}.

A main intuition behind probing is to evaluate to what degree 
the representations that are learned from the complex source task can be re-purposed to solve a new, simpler task \cite{belinkov-2022-probing}. In our work we decide to reverse this paradigm (thus \textit{reverse probing}) and investigate how different source task embeddings, from a model trained on simple tasks, can be adapted to a new, more complex target task. In other words, we try to answer the question: \textbf{Can we reuse knowledge from simpler tasks for a more complex task?} Such a \textit{recycling} of knowledge is not only interesting to deepen our understanding of what type of knowledge is encoded in language models, but it also results in more energy-efficient deep learning, by reusing network weights and representations.

We choose \textbf{coreference resolution} \cite{lee-etal-2017-end} as our target task because solving coreference is - up to date - a challenging NLP problem that even newer large language models struggle with \cite{10.1162/tacl_a_00543,martinelli-etal-2024-maverick}. Coreference resolution involves understanding of context, what counts as a valid mention and which mentions refer to the same entity. Solving coreference may require different types of linguistic knowledge. Our goal is to find out which types of information from which source task models are useful and how this information can be combined and/or adapted to work for the target task.

To isolate the effects of single tasks, we rely on small language models, in our case BERT \cite{DBLP:conf/naacl/DevlinCLT19}. Such models do not possess sophisticated in-context abilities and require finetuning steps in order to perform well on tasks. Our research questions are as follows:

(1) Which \textbf{source tasks} are beneficial for combination into a more complex target task, here coreference resolution? 

(2) Which \textbf{layers} of source task models contribute most to the target model performance?

(3) What are the effects of \textbf{combining embeddings} from different models and layers? How should these embeddings be aggregated? Can we improve word representations by extending the \textbf{embedding context} and combining the outputs of several hidden layers?

\section{Reverse Probing} \label{sec:architecture}
The goal of our framework is to evaluate the transferability of knowledge embedded in representations from simpler source tasks to a complex target task. Figure \ref{fig:probing-workflow} gives an overview.

Let \( S = \{s_1, s_2, \dots, s_k\} \) be a set of source tasks with models pre-trained on simpler NLP tasks, and \( T \) be the target task (coreference resolution in this case). \( M_s \) is a pre-trained model fine-tuned on source task \( s \). We then define \( H^l_s \) to be the output embeddings from layer \( l \) of \( M_s \). 

For each source task \( s_i \in S \), we extract embeddings \( \mathbf{H}_s \) from layer \( l \) of the corresponding source task model \( M_s \) (Figure \ref{fig:probing-workflow}, embedding extraction). We either take the output at a single or multiple consecutive layers. Note that these may be also intermediate layers (model truncation). Optionally, we apply L$_2$ normalization.

Secondly, we aggregate token embeddings from different source task models by using an aggregation function \( A \) to combine embeddings across layers and models. The aggregation is done token-wise, so that every token can be represented as a combination of different model outputs. 
We define \(A\) to be either the mean of all vectors, i.e. as 
\[
\mathbf{E}_T = \frac{1}{k} \sum_{i=1}^k \mathbf{H}_{s_i}
\]
Alternative, we use a simple attention mechanism for the combination (Figure \ref{fig:probing-workflow}, embedding aggregation):
\[
 \mathbf{E}_T = \sum_{i=1}^k \alpha_i \mathbf{H}_{s_i}
 \]
 where 
\[
\alpha_i = \text{softmax}(\mathbf{W} \cdot \mathbf{H}_{s_i})
\]
         
In some experiments we use only a single model. In this case the mean corresponds to the original embedding of the source model and attention simply means self-attention.

Next, the aggregated token embeddings are passed to the target task head that includes several layers with trainable weights (Figure \ref{fig:probing-workflow}, training target task layers). 

Figure \ref{fig:probing-workflow} shows the probing workflow with four different source task models. Each source model is pre-trained separately on a corresponding dataset as described in Section \ref{sec:data-tasks}. The models are based on \textit{bert-base-cased} contextualized embeddings with different task-specific heads and their weights are frozen. Given that the source models cannot update their weights during probing, our conjecture is that those models that perform better on the coreference task ``out-of-the-box" contain some useful information that is relevant for the target task.

\section{Tasks}\label{sec:data-tasks}

In this section we describe the target task, the source tasks and their respective training data. 

\subsection{Target Task}

As our target model we choose a popular end-to-end coreference resolution model based on the implementation by \cite{xu-choi-2020-revealing} and train it on the OntoNotes CoNLL 2012 corpus \cite{pradhan-etal-2012-conll}. We use \textit{bert-base-cased} and the recommended parameters for fine-tuning\footnote{\url{https://github.com/lxucs/coref-hoi/blob/master/experiments.conf}}. 

\subsection{Source Tasks}

We focus on the comparison with standard BERT as well as four other task-specific models. As source tasks we take a range of tasks of varying complexity: Paraphrase identification, named entity recognition, relation extraction, and - a (more complex) source task - quesion answering. 

The first model is fine-tuned on the Microsoft Research Paraphrase Corpus (MRPC) \cite{dolan-brockett-2005-automatically}. Since paraphrased sentences describe the same entities and events, such sentence pairs likely contain more coreferent mentions than standard (non-paraphrased) texts. Hence, MRPC embeddings are more tuned towards semantic similarity and could be useful for the coreference task.

Named Entity Recognition (NER) model is trained on the CoNLL 2012 dataset \cite{pradhan-etal-2012-conll} and can generate one of the 37 labels for each token (e.g., PERSON, PRODUCT, DATE etc.). Named entities are often involved in coreference relations and being able to identify mention spans correctly is crucial for coreference resolution.

Next, we also experiment with the Relation Extraction model (RE) trained on the TACRED dataset \cite{zhang-etal-2017-position}. It provides annotations for the spans of the subject and object mentions as well as the mention types according to the Stanford NER system and relations (if applicable) between the entities. Similarly to the NER model, RE is important for coreference because one of the tasks that this model performs is mention span detection. However, it also classifies different relations between the mentions and such relations are typically non-referential (e.g. \textit{``per:employee\_of"}).

Another source task model used in this project is trained on the SQUAD 2.0 dataset \cite{rajpurkar-etal-2016-squad} for extractive question answering. This model (QA) can identify answer spans given the question and a paragraph of text. Since answering questions often involves coreference resolution, there is an overlap between these two tasks and word embeddings from one task might be beneficial for another.

For single model experiments we also analyse the performance on the vanilla BERT model \footnote{We use the cased variant from HuggingFace under \url{https://huggingface.co/bert-base-cased}.}\cite{DBLP:conf/naacl/DevlinCLT19} which was trained with a masked language modeling objective on BookCorpus \cite{DBLP:conf/iccv/ZhuKZSUTF15} and English Wikipedia. 
Note that all the other source models are fine-tuned versions of this model.

Additionally, we experiment with the POS-tagging model\footnote{\url{https://huggingface.co/QCRI/bert-base-cased-pos}}, the models for semantic tagging\footnote{\url{https://huggingface.co/QCRI/bert-base-cased-sem}} and chunking\footnote{\url{https://huggingface.co/QCRI/bert-base-cased-chunking}} as well as another NER model (NER-dslim)\footnote{\url{https://huggingface.co/dslim/bert-base-NER}} trained on the English version of the CoNLL-2003 Named Entity Recognition dataset \cite{tjong-kim-sang-de-meulder-2003-introduction}. However, we limit the number of experiments for these models and focus mostly on MRPC, NER, RE and QA tasks.

\section{Experiments and Results}\label{experiments}

In this section we describe our experiments with various source models and probe them on the coreference resolution task (\S\ref{subsec:source-task-choice}). We also evaluate different embedding aggregation methods (\S\ref{subsec:aggregation-strategies}), measure the effects of using intermediate layer output and normalization (\S\ref{subsec:truncation-norm}), vary the embedding context from several hidden layers (\S\ref{subsec:embedding-context}) and compare the performance of multiple vs single models (\S\ref{subsec:one-vs-many}).

\subsection{Training Details and Evaluation}

The coreference-specific layers are trained with the learning rate 1e-4 and early stopping (maximum number of epochs is set to 100 and patience is set to 5). The learning rate was optimized based on the experiments with the standard frozen BERT model.

For evaluation we use an average F1 score that is a combination of MUC \cite{vilain-etal-1995-model}, CEAF \cite{luo-2005-coreference} and B$^3$ \cite{bagga-baldwin-1998-entity} coreference metrics. We run each experiment with three different seeds and report the average F1 values on the validation set. The target (non-frozen) model trained on the coreference resolution task from scratch achieves 73.75 F1 which is an upper bound for our probing task.

\subsection{The Choice of the Source Task Models}\label{subsec:source-task-choice}

\begin{figure*}[t]
    \centering
    \includegraphics[width=.7\textwidth]{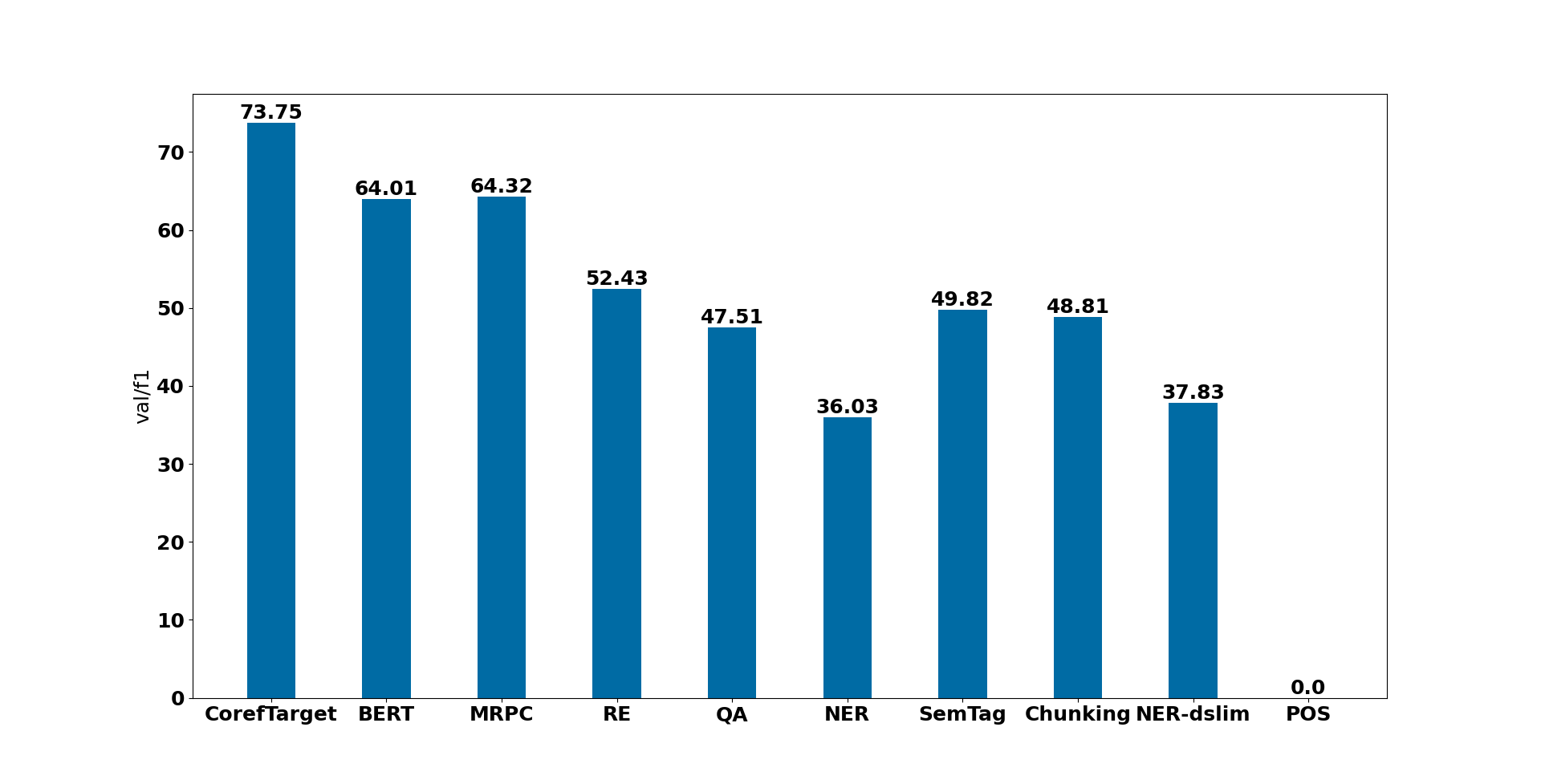}
    \caption{Source task models: CorefTarget, BERT, MRPC, RE, QA, NER, SemTag, Chunking, NER-dslim, POS}
    \label{fig:source-tasks}
\end{figure*}

Figure \ref{fig:source-tasks} shows the comparison between different source task models. Our original set of models that includes MRPC, NER, RE, QA and vanilla BERT has two clear winners: BERT and MRPC (64.01 and 64.32 F1). They are followed by RE (52.43) and QA (47.51) models and, finally, NER achieves the lowest score of 36.03. This comparison is based on a single run with the same seed, the averaged results across three runs with standard deviation can be found in Table \ref{tab:single-models}.

\begin{figure*}[t]
    \centering
    \includegraphics[width=.7\textwidth]{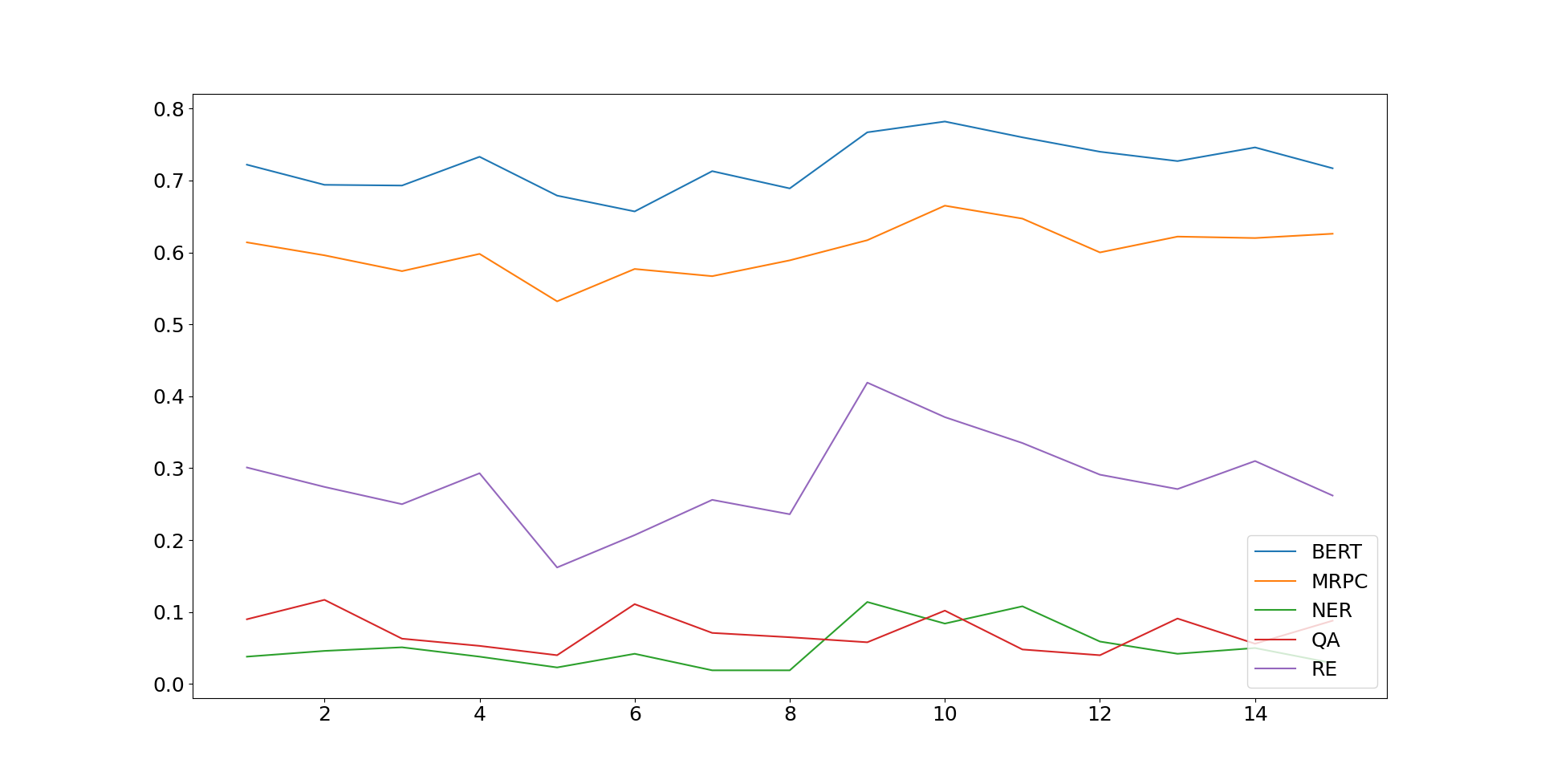}
    \caption{Average cosine similarity between the embeddings of the source tasks and the target coreference task, averaged across all tokens for 15 batches}
    \label{fig:cossim-embed}
\end{figure*}
\input{tables/with_variance/single_models}
\input{tables/with_variance/2x_models}
We also have a closer look at the cosine similarity between our source models and the pre-trained coreference model. Figure \ref{fig:cossim-embed} shows similarity scores averaged across all tokens for 15 random batches. The scores are collected before the embedding aggregation. Hence, they show how close the original source model embeddings are to the ``ideal" task embeddings. Unsurprisingly, BERT and MRPC have the most similar embeddings to the coreference target. On the other hand, although both QA and NER embeddings are very different from the target task embeddings, QA achieves much better performance than NER on this task (50.79 vs 35.63 F1, see Table \ref{tab:single-models}). This shows that even though cosine similarity is a good approximation for the task similarity, it is not an ideal predictor for the target task performance and even the source models with very different embeddings (QA) can still achieve the scores comparable to the ones achieved by the models with more similar embeddings (RE).

Additional models that we tested demonstrate rather poor performance on the coreference resolution task (see Figure \ref{fig:source-tasks}). POS-tagging model struggles to learn anything about coreference and the training does not progress. Another NER model trained on a different version of Ontonotes (NER-dslim) achieves the maximum of 37.83 F1. Chunking and semantic labeling are somewhat more successful and achieve 48.81 and 49.82 F1 each, correspondingly.

\subsection{How to Combine Task Embeddings}\label{subsec:aggregation-strategies}

We employ two different aggregation strategies to combine the embeddings of the source task models: mean and attention-based aggregation. Additionally, we experimented with summing instead of using the mean, but the results were comparable or slightly worse: A combination of frozen MRPC with BERT achieves 62.34 F1 with sum and 63.26 with mean (average values across three runs with different seeds). Hence, in all further experiments we focus on the comparison between the mean and attention-based aggregation.

\input{tables/with_variance/multiple_models}

\begin{figure*}[t]
\centering
\begin{subfigure}{.49\linewidth}
\centering
\includegraphics[scale=0.18]{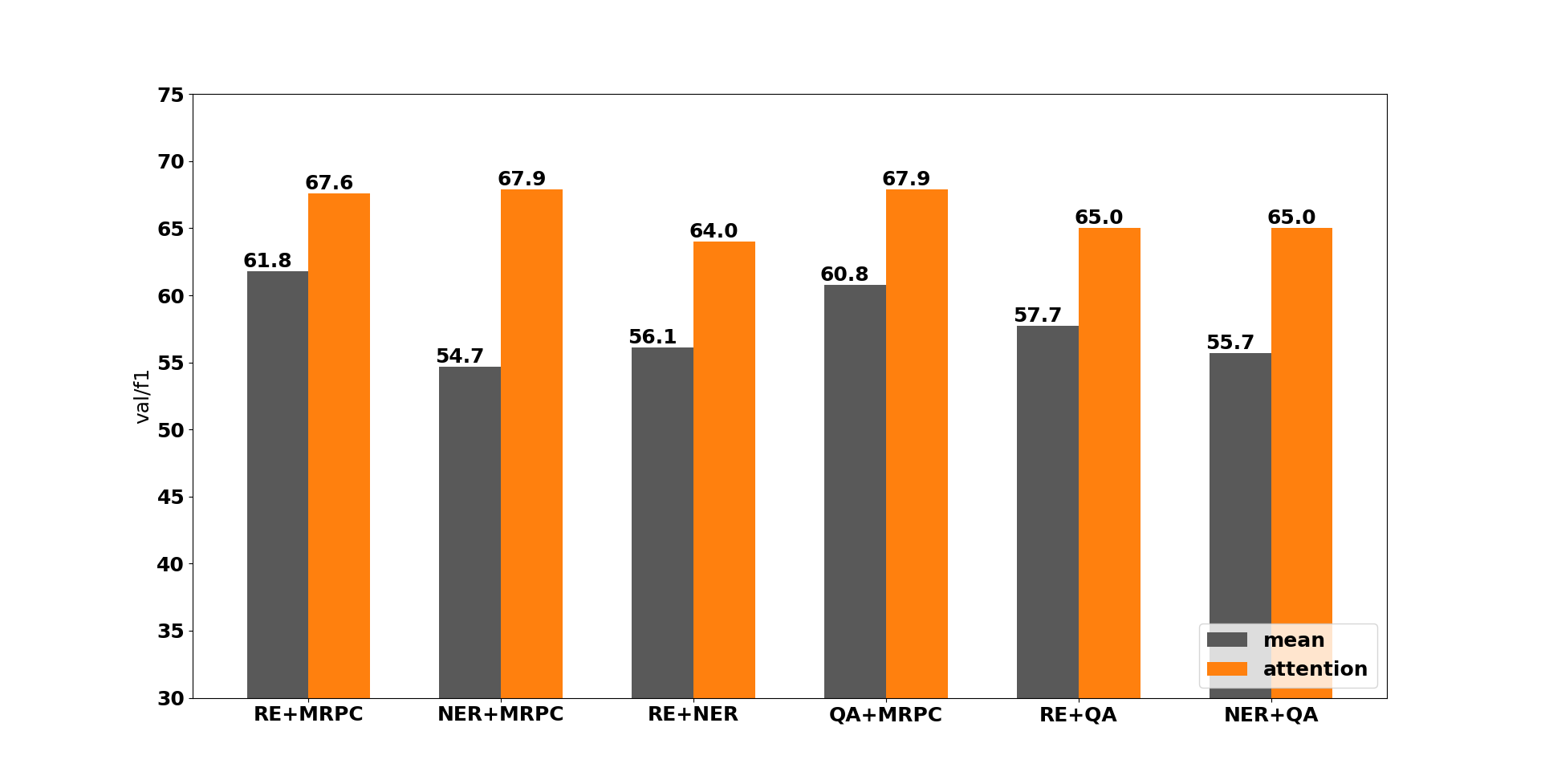}
\caption{Pairs of models: comparison of two embedding aggregation methods, mean and attention, to combine the source task model outputs}
\end{subfigure}
\hfill
\begin{subfigure}{.49\linewidth}
\centering
\includegraphics[scale=0.18]{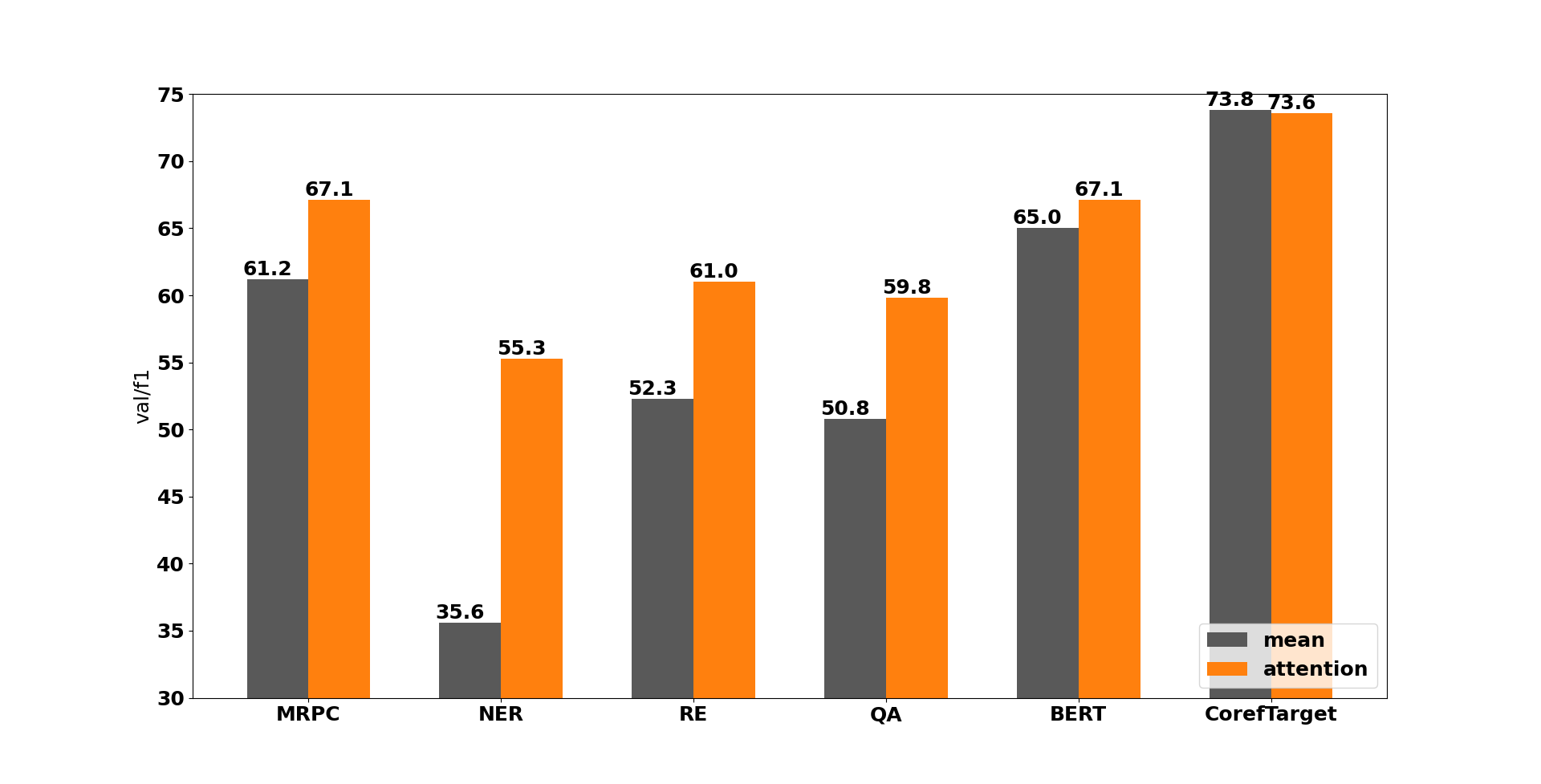}
\caption{Single models: 
performance gains by adding attention projections (attention) compared to having no additional parameters (mean) 
}
\end{subfigure}%
\caption{Mean vs attention aggregation (full setting)}\label{fig:attention-vs-mean}
\end{figure*}

F1 scores for single models as well as for their 2x, 3x and 4x combinations can be found in Tables \ref{tab:single-models}, \ref{tab:2x-models} and \ref{tab:multiple-models}. We also summarize the results for single models and for pairs of models graphically in Figure \ref{fig:attention-vs-mean} that shows how much models benefit from attention. However, this trend holds even when there is only a single source model. This shows how much improvement we get by simply adding additional projections in the case of attention aggregation. The performance gains are different depending on the model. E.g., if we use pre-trained coreference model as our source task, there is almost no difference between attention and mean aggregation. However, other task-specific models can substantially benefit from selective aggregation. E.g., NER gains almost +19.7\% and QA improves by +9\%. In general, all models except for the one that has the same source and target tasks (CorefTarget) benefit from attention and improvements are larger for the models that have lower original scores.

For multiple model combinations we also see a similar trend with consistent improvements when attention-based aggregation is used, e.g., +13.23\% for NER+MRPC and +9.36\% for NER+QA (see Figure \ref{fig:attention-vs-mean} and Table \ref{tab:2x-models} for further comparisons). Interestingly, when we combine our source models with the model that was pre-trained on the coreference task (CorefTarget), we have only negligible improvements because the attention aggregator quickly learns which source model is beneficial for the task and starts paying almost all attention to the output of this model ignoring all the others. However, if we do not add coreference task to the set of source tasks we observe some interesting patterns that emerge with the combinations of different models. Figure \ref{fig:attention-mrpc-ner-re} (in the Appendix) shows how attention is distributed across different training epochs for the combination of MRPC, RE and NER. In the beginning, all three models are being paid the same amount of attention ($\approx$33\%). However, the aggregator soon starts prioritizing MRPC and NER gets progressively less and less attention. Interestingly, RE model also loses some impact over time but more slowly and remains somewhat important for the aggregator until the end of the training.

\subsection{How to Extract Embeddings}\label{subsec:truncation-norm}

\begin{figure*}[t]
    \centering
    \includegraphics[scale=.25]{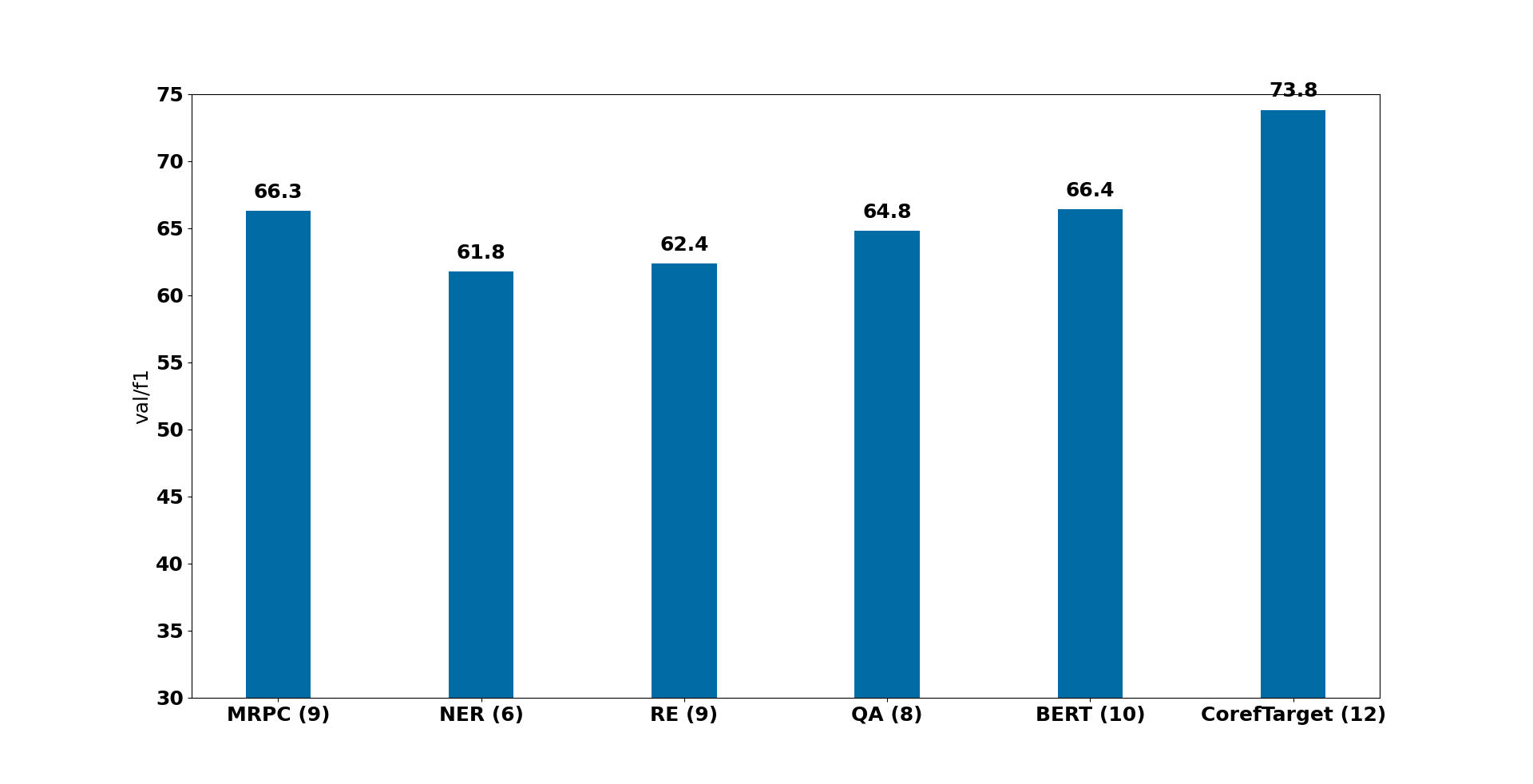}
    \caption{Source task model performance truncated to the best layer (in parentheses) with mean aggregation}
    \label{fig:mean-best-truncated-layer}
\end{figure*}

\begin{figure*}[t]
\centering
\begin{subfigure}{.49\linewidth}
\centering
\includegraphics[scale=.17]{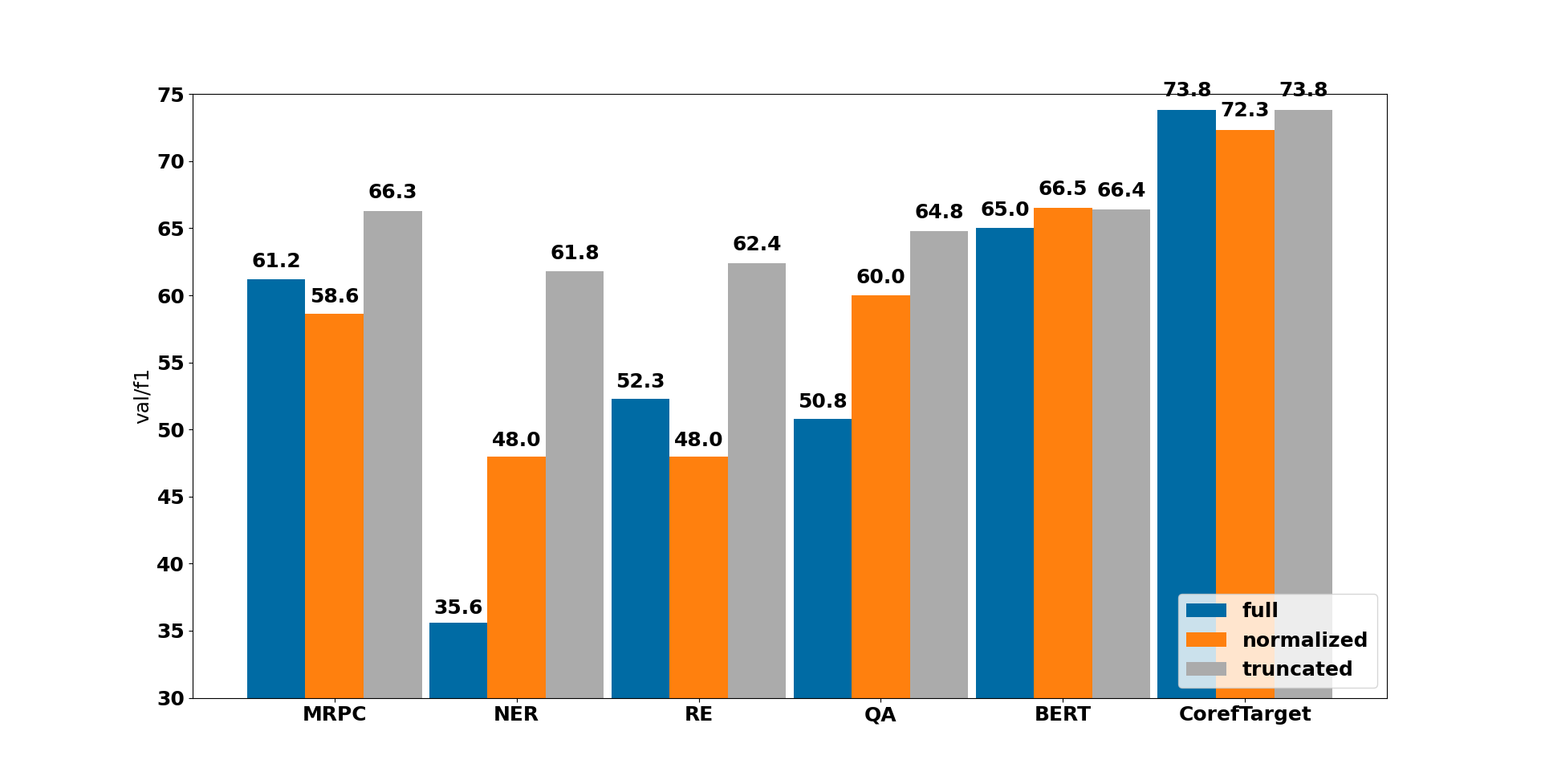}
\caption{Mean aggregation}
\end{subfigure}
\hfill
\begin{subfigure}{.49\linewidth}
\centering
\includegraphics[scale=.17]{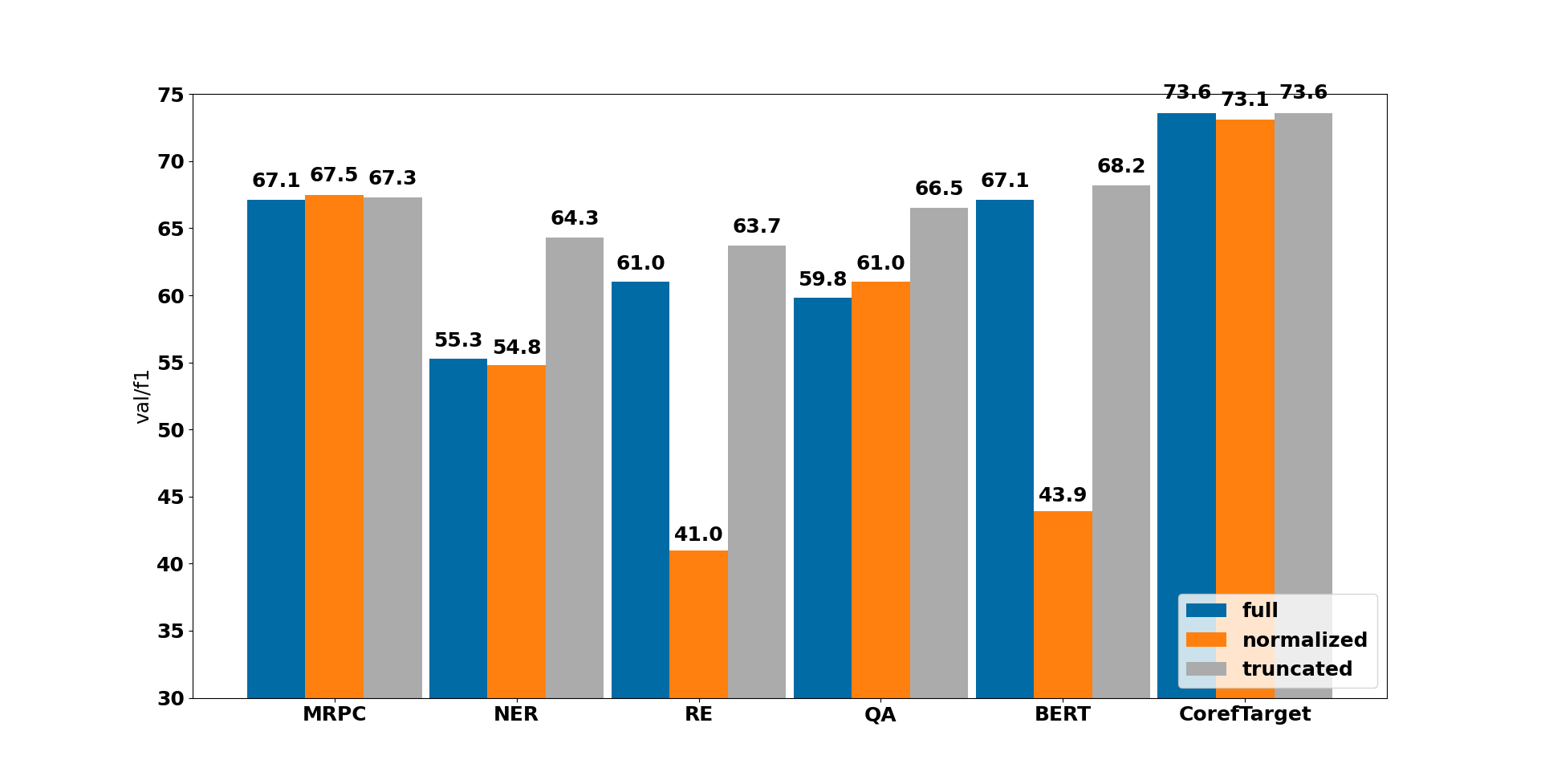}
\caption{Attention aggregation}
\end{subfigure}%
\caption{Single models: full vs normalized vs truncated}\label{fig:single-models-mean-vs-attn-aggregation}
\end{figure*}

\begin{figure*}[t]
\centering
\begin{subfigure}{.49\linewidth}
\centering
\includegraphics[scale=.17]{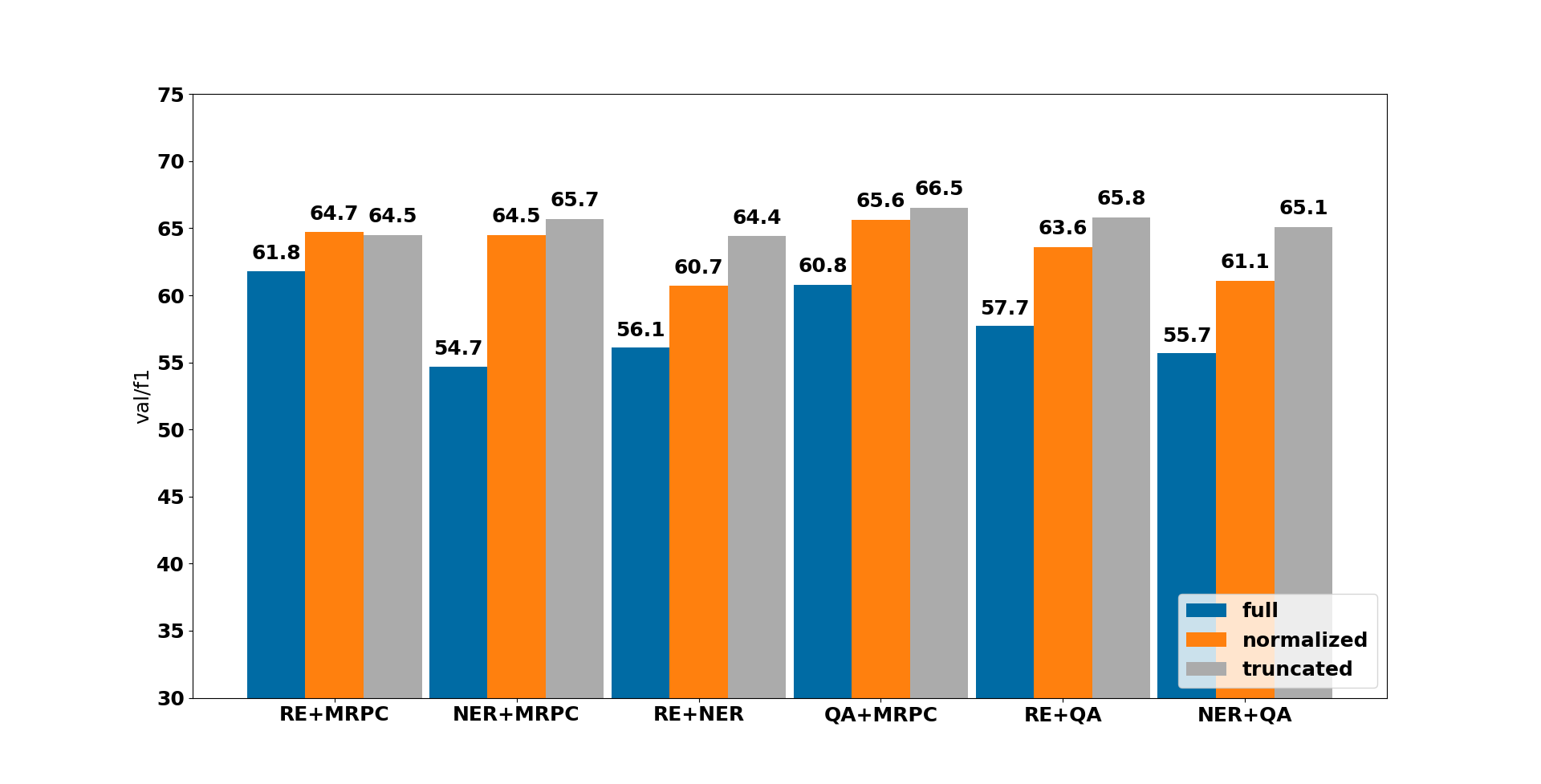}
\caption{Mean aggregation}
\end{subfigure}
\hfill
\begin{subfigure}{.49\linewidth}
\centering
\includegraphics[scale=.17]{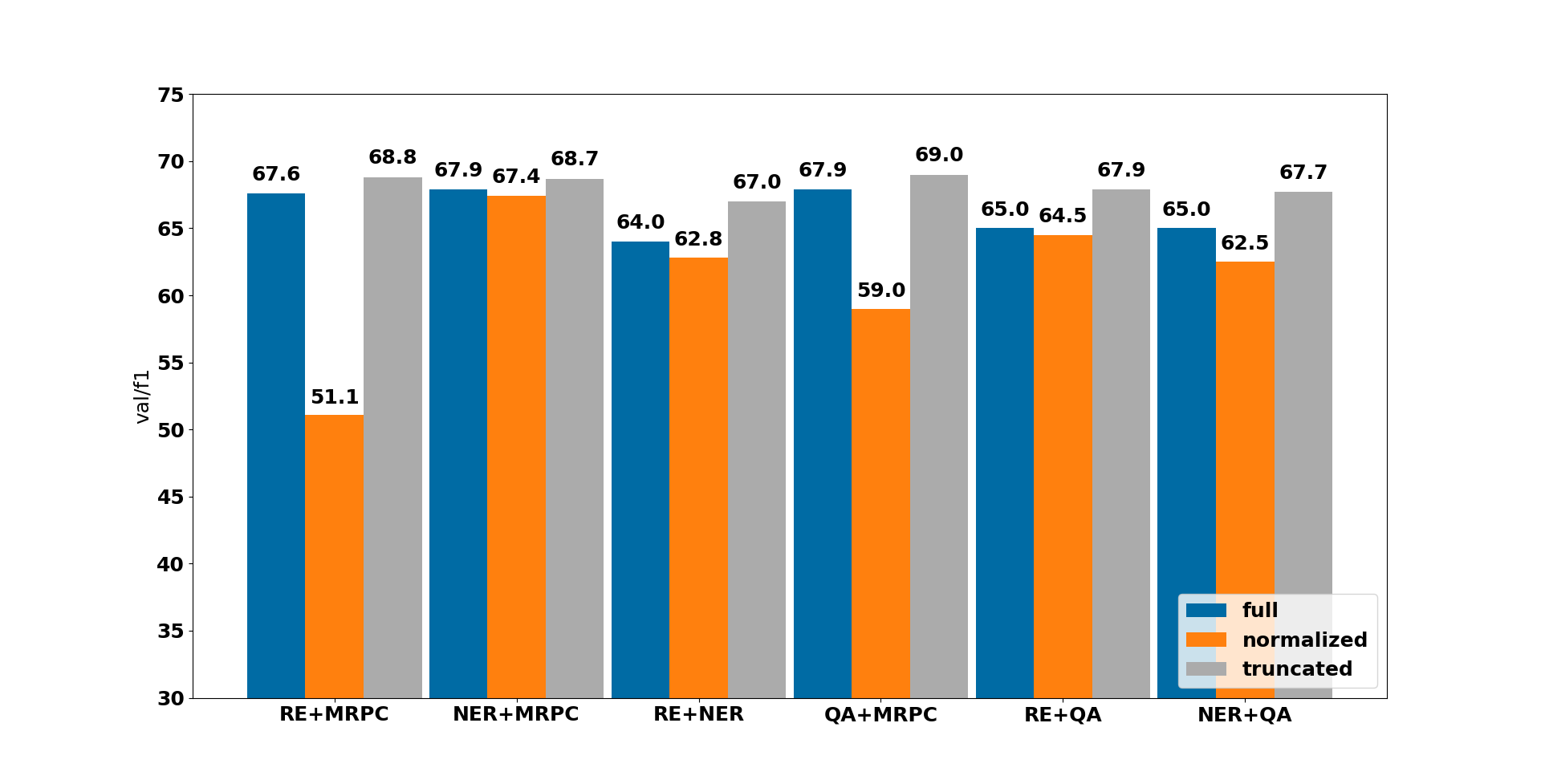}
\caption{Attention aggregation}
\end{subfigure}
\caption{Pairs of models: full vs normalized vs truncated}\label{fig:2x-models-mean-vs-attn-aggregation}
\end{figure*}

We also consider different ways of embedding manipulations since the final layers of BERT-based models might be too specialized on their corresponding tasks, so that their representations are no longer useful for coreference resolution. In fact, after comparing the embeddings from layer 6 to 12 we found that the best performing layer on our probing task was typically not the final one. E.g., it was layer 9 for MRPC and RE, layer 8 for QA and 6 for NER (see Figure \ref{fig:mean-best-truncated-layer}). Tables \ref{tab:single-models}, \ref{tab:2x-models} and \ref{tab:multiple-models} show the detailed comparisons between the original (full) model outputs as well as the normalized and truncated (to the ``best" layer) versions for single models and their combinations (see also Figure \ref{fig:single-models-mean-vs-attn-aggregation} and \ref{fig:2x-models-mean-vs-attn-aggregation} for the plot comparison). Truncation seems to be a good strategy for embedding aggregation and consistently yields best results across different settings. Truncation improves NER by up to +26.2\%. QA is improved by +14\% (Figure \ref{fig:single-models-mean-vs-attn-aggregation}). Also combinations of models work better with truncation, e.g., RE+QA pair gains +8.17\% F1 with mean aggregation and +2.84\% with attention aggregation when both models are truncated (Figure \ref{fig:2x-models-mean-vs-attn-aggregation}).

Since combining embeddings from disparate models is a challenging task, especially when the source tasks are quite different from each other, we also experiment with applying L$_2$ norm to the output of each model before aggregating the embeddings. This gives us varied results depending on the model and the aggregation type. E.g., for mean aggregation single NER, QA and BERT benefit from normalization but MRPC and RE result in lower scores. For attention aggregation all models except for MRPC and QA have substantial drop in performance.

It is also interesting to see the effect of normalization on the combinations of different models. For mean aggregation normalization brings substantial improvements, e.g., +9.76 F1 for NE+MRPC and +5.48\% for NER+QA and, in general, all 2x models show better performance with normalization (Table \ref{tab:2x-models}). However, there is a very different trend for attention-based aggregation. Here we see a large drop in performance for most of the models, e.g., -8.87 F1 for QA+MRPC which indicates that attention can already combine the embeddings from different models quite well and achieves worse results with more uniform normalized embeddings.

\subsection{Embedding Context from Multiple Layers}\label{subsec:embedding-context}

Since we found in our previous experiments that truncation consistently improves the performance for many source models, we decided to explore another setting that involves concatenating the embeddings of the last n hidden layers of the source model before aggregating them with attention. We experiment with the last 4, 6 and 12 layers and compare them to the aggregation that affects only the last layer of each model (see Table \ref{tab:single-models} for more detail).

Our results show that for single models having more ``embedding context" is beneficial. Overall, combinations of the last 4 or 6 layers result in the best F1 scores. However, combining all layers of the model is not necessarily useful and can even hurt the performance. E.g., NER achieves 66.30 F1 with combined 6 layers which is +11 F1 improvement compared to the same model that uses only a single last layer but when we combine all 12 layers of NER the metric decreases from 66.30 to 65.76 F1 (Table \ref{tab:single-models}).

Another interesting observation is that for vanilla BERT combining the outputs of the last 4 or 6 layers does not make any difference, and for other models the difference is more pronounced, although still rather small. NER is the model that gains the most from the increased embedding context, it gains additional +2.5 F1 by combining 6 instead of 4 last layers which is also consistent with our finding for the truncated models and the fact that NER performs better when truncated to 6 layers. The only model that does not show any improvements in the layer concatenation setting is the coreference source model since it is already optimized for the task and performs best as it is, i.e., without truncation, normalization or any other embedding manipulations.

\subsection{Combining Multiple Source Models}\label{subsec:one-vs-many}

An interesting research question with respect to the embedding aggregation is how many models are actually needed to achieve good results and whether such models should be more or less similar to each other. E.g., NER and RE both focus on mention span extraction, RE and QA process relations between the entities in the text and MRPC model is more suitable for the semantic similarity tasks.

\begin{figure*}[t]
\centering
\begin{subfigure}{.49\linewidth}
\centering
\includegraphics[scale=.17]{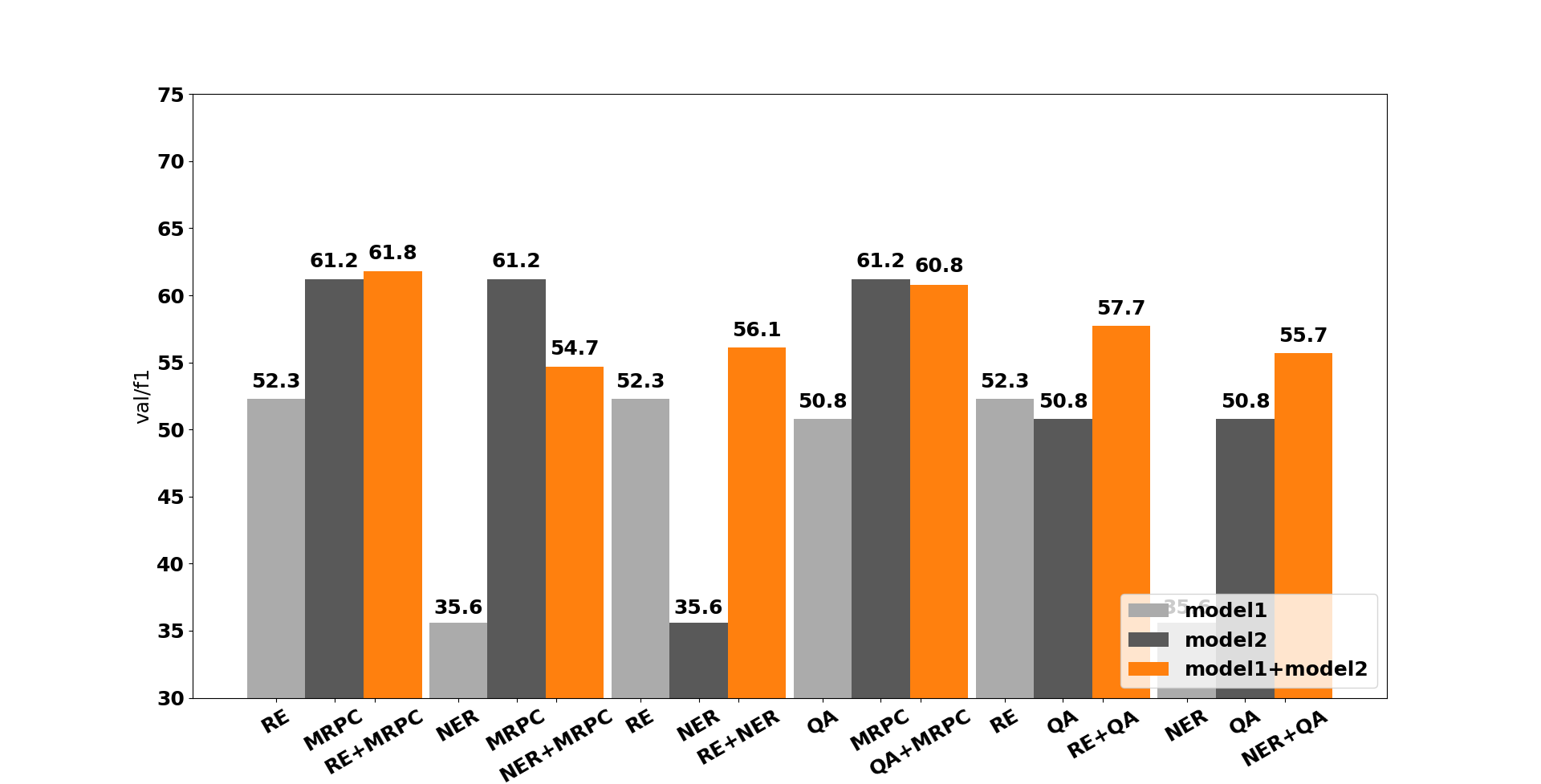}
\caption{Mean aggregation}
\end{subfigure}
\hfill
\begin{subfigure}{.49\linewidth}
\centering
\includegraphics[scale=.17]{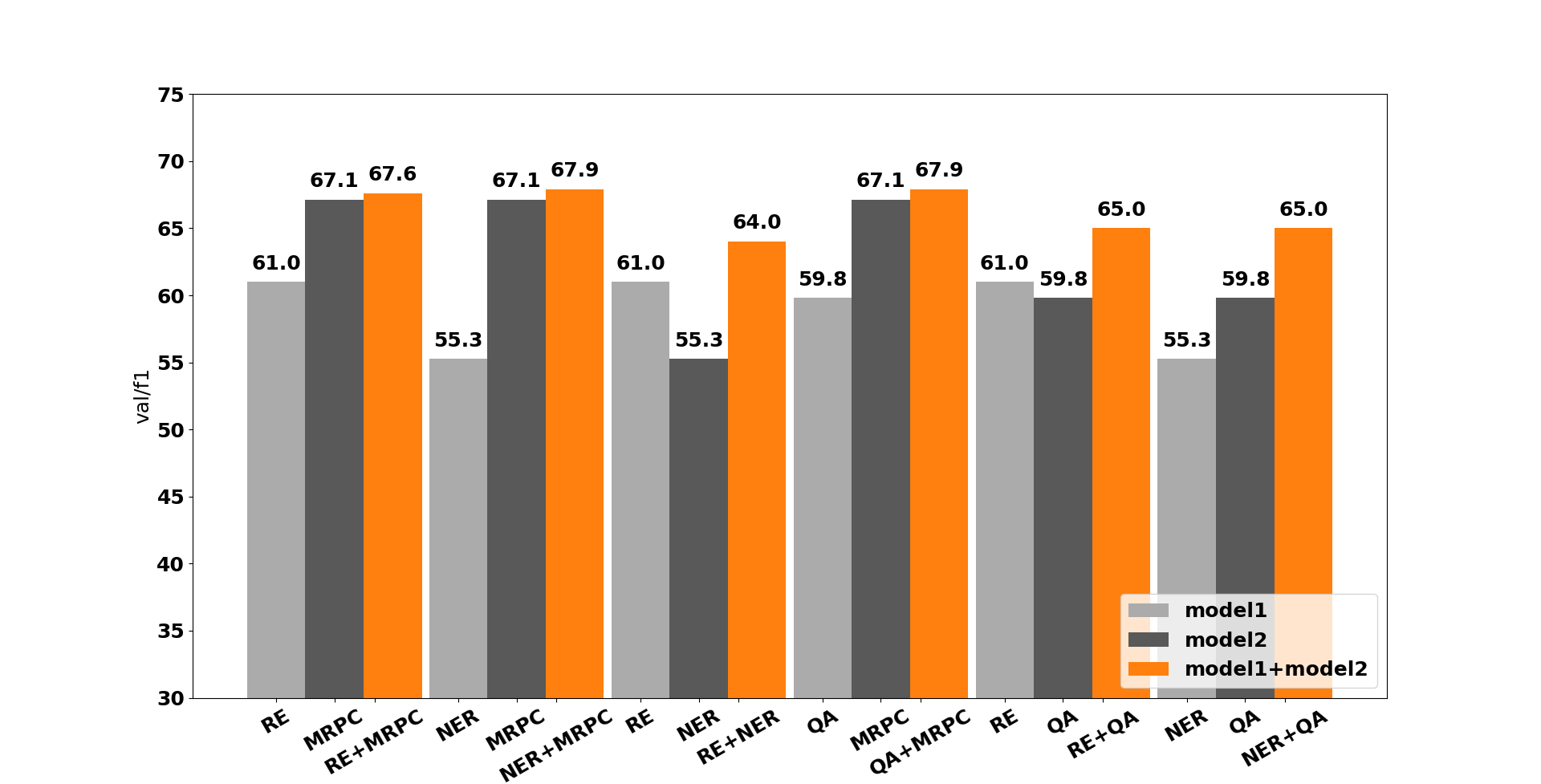}
\caption{Attention aggregation}
\end{subfigure}%
\caption{Single and combined (2x) models}\label{fig:single-vs-2x}
\end{figure*}

Firstly, we found that combinations of two models always outperform single models in the attention aggregation setting and, for the mean setting, pairs of models typically also perform better than the individual models except for the combinations with MRPC that tend to have lower scores (see Figure \ref{fig:single-vs-2x} for the comparison with mean and attention). E.g., NER with mean aggregation achieves 35.6 F1, RE achieves 52.3 and the combination of both (RE+NER) has 56.1 F1.

Secondly, we observed that combining three or more models typically works well for the full models. However, for the truncated setting there are only negligible gains when we combine multiple models (e.g., for RE+MRPC with attention we have 68.78 and for RE+MRPC+QA 68.81).

Lastly, model combinations that include MRPC tend to perform better than the rest which likely indicates the importance of semantic similarity for the coreference task. However, the combinations of RE+NER, RE+QA and QA+NER can also be beneficial, especially in the mean aggregation setting.

\section{Related Work}\label{sec:related-work}
Apart from the work on probing that was presented in the introduction, our work is closely related to the idea of \textit{transfer learning} \cite{torrey2010transfer}, one of the ubiquitous paradigms in modern NLP. The idea of transfer learning is to train a model on a task A and then transfer the weights to a task B, either with or without further finetuning. This is the basis of most modern language models, which are pretrained and then applied or evaluated on specific downstream tasks \cite{DBLP:conf/naacl/DevlinCLT19,raffel2020exploring,jiang2023mistral,dubey2024llama}.
The pretraining data of large language models often contains samples from various natural language tasks, which renders most language models as multi-task learners \cite{yu2024unleashing}. Multi-task learning describes a paradigm where a model is simultaneously trained on a range of tasks. While this concept is related to the work presented here, the main difference is that several source tasks are mixed together during (pre-)training usually, which is not the case with our work. 

In some sense, our work is related to research around task arithmetics \cite{NEURIPS2022_70c26937,chronopoulou2023language,ilharco2023editing,belanec2024task}, which has the goal to explicitly compute task representations in networks, e.g. as differences to a random initalization, and implement transfer learning by means of difference vector arithmetics. In contrast, our work concentrates on hidden representations, rather than the parameters of the network.

\section{Conclusion}\label{sec:conclusion}

In this project we ``reversed'' the classical probing and investigated how different source task embeddings contribute to a target task (coreference resolution). Our experiments with Paraphrase Detection (MRPC), Named Entity Recognition (NER), Relation Extraction (RE) and Extractive Question Answering (QA) as source tasks show there are quite different embedding representations that achieve different scores on the target task ranging from 35.63 F1 (NER) to 61.16 F1 (MRPC) for single models. 

Moreover, we found that the best performing embeddings were typically not the outputs of the last hidden layer but rather the representations generated at lower layers of the embedding model. MRPC was found to to be the best source model, while, surprisingly, NER performed worst. 

We also explored different combinations of source models and found that two or more models typically outperform single ones. We considered mean and attention-based embedding aggregation methods and demonstrated the effectiveness of attention. For single models, we also showed that combining the outputs of several hidden layers instead of only one layer is beneficial. However, combining the outputs of all available layers is not necessarily a good strategy and usually the best scores can be achieved by combining only the outputs of the last 4 hidden layers that possibly contain more high-level, semantic information important for the coreference task.

In the future it would be interesting to experiment with more types of embedding manipulations. Also, a combination of truncation and normalization could possibly bring some gains for single models. Moreover, it would be interesting to check the effects of attention aggregation with hidden layer concatenations for multiple models (e.g., RE+MRPC). So far we tested the approach that proved to be successful only for single models. Finally, it would be interesting to replicate our experiments on larger (non-BERT) models and tasks (e.g., semantic role labeling, discourse relation classification etc.).

We hope that our experiments can help to clarify the impact of embeddings and their combinations on the target coreference task. We also hope that the reverse probing idea will facilitate further research on finding useful information in the task-specific representations that originate from different fine-tuned models.

\section*{Limitations}
While this work sheds light on the potential of reverse probing and task embeddings, some limitations arise.

First, we exclusively work with BERT-based models. This gives us a controlled setup, but it also means our findings might not fully translate to largermodels or other architectures like GPT, T5, or multilingual models. Future work needs to investigate a broader range of models.

Our choice of source tasks Paraphrase Detection (MRPC), Named Entity Recognition (NER), Relation Extraction (RE), and Question Answering (QA) - is not exhaustive. There are many other NLP tasks, such as sentiment analysis, syntactic parsing, or commonsense reasoning, that might contribute useful embeddings for coreference resolution. Also, some of the tasks are not necessary simpler tha coreference resolution (e.g., QA), which we pickedas our target task. Generally, our conclusions are centered around coreference resolution as the target task. While this is a challenging and linguistically complex problem, our approach may not directly apply to other NLP tasks with different dependencies, such as machine translation or text summarization. Testing on additional target tasks would be a logical next step.

Lastly, there’s the question of computational efficiency. While we worked with relatively small models, combining embeddings from multiple layers and tasks does introduce extra processing overhead. Finding ways to leverage task embeddings efficiently, without significantly increasing inference costs, is another area worth exploring. 

\section*{Acknowledgments}
This work has been supported by the German Ministry of Education and Research (BMBF) as part of the project TRAILS (01IW24005).

\bibliography{acllatex}
\newpage
\appendix
\section{Additional Figures}
\begin{figure}[h]
\centering
\begin{subfigure}{\linewidth}
\centering
\includegraphics[scale=.18]{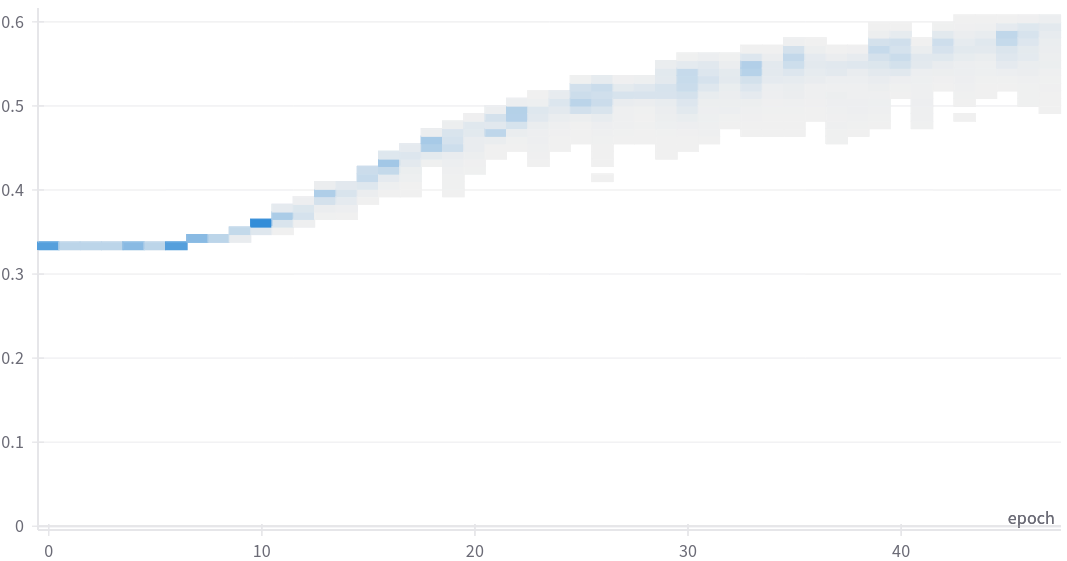}
\caption{MRPC}
\end{subfigure}
\vfill
\begin{subfigure}{\linewidth}
\centering
\includegraphics[scale=.18]{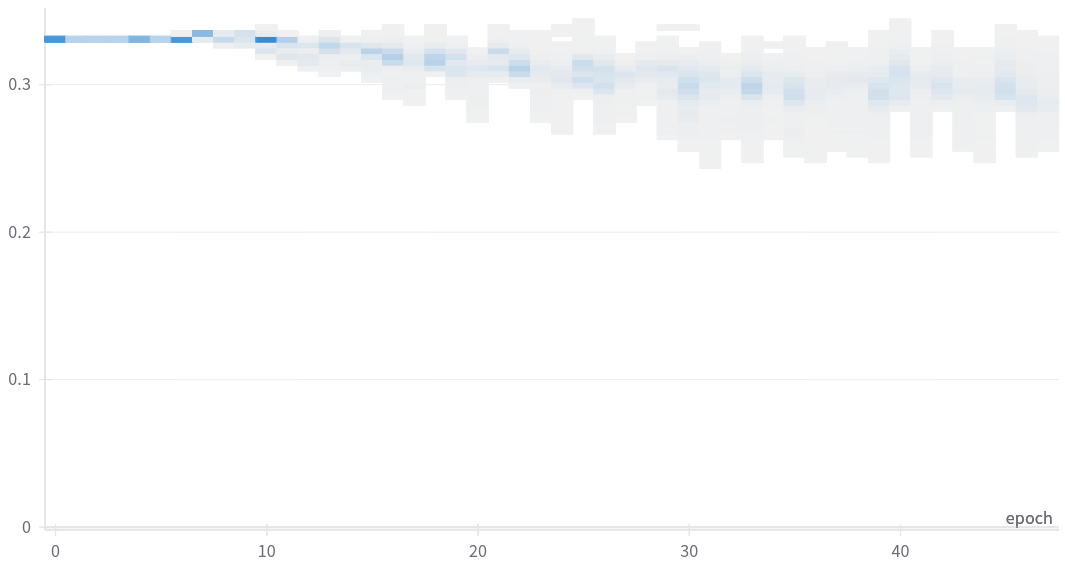}
\caption{RE}
\end{subfigure}
\vfill
\begin{subfigure}{\linewidth}
\centering
\includegraphics[scale=.18]{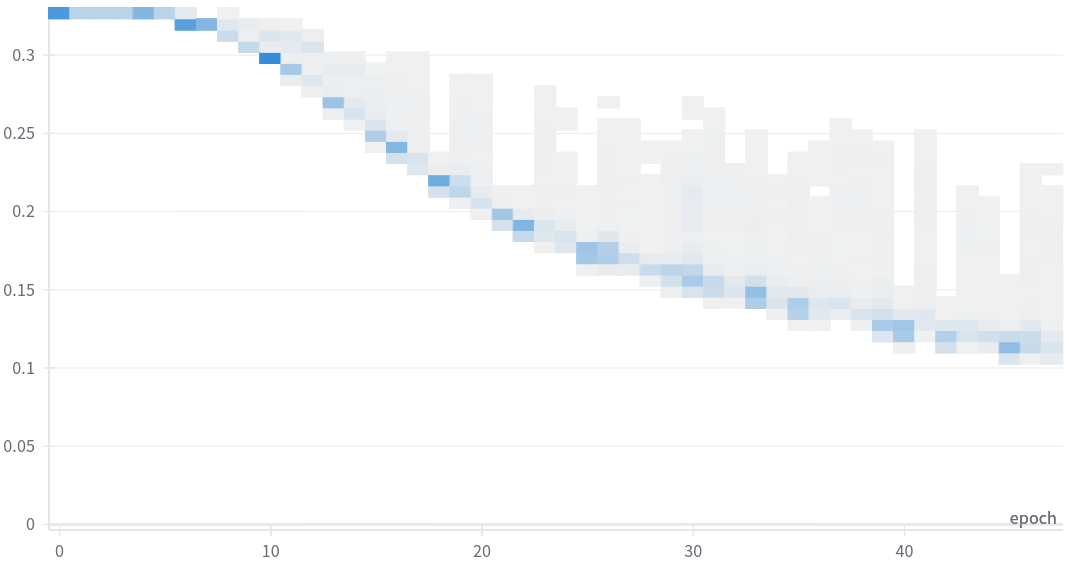}
\caption{NER}
\end{subfigure}%
\caption{MRPC+RE+NER with attention aggregation (full setting)}\label{fig:attention-mrpc-ner-re}
\end{figure}



\end{document}

%% file: tables/with_variance/single_models.tex
\begin{table*}[t]
\centering
\scalebox{0.75}{
\begin{tabular}{|l|l|l|l|l|l|l|l|l|l|}
\hline
\multirow{2}*{models} & \multicolumn{3}{|c|}{mean} & \multicolumn{3}{|c|}{attention} & \multicolumn{3}{|c|}{layer concat + attention} \\
\cline{2-10}
& full & norm & trunc & full & norm & trunc & 4 & 6 & 12  \\
 \hline
 MRPC$_{(9)}$ & $61.16_{\pm2.84}$ & $58.61_{\pm{13.7}}$ & $66.26_{\pm0.61}$ & $67.05_{\pm0.70}$ & $67.49_{\pm0.29}$ & $67.27_{\pm0.30}$ & $\textbf{67.94}_{\pm1.54}$ & $67.03_{\pm0.99}$ & $67.28_{\pm1.48}$ \\
 NER$_{(6)}$ & $35.63_{\pm2.12}$ & $47.95_{\pm3.27}$ & $61.76_{\pm1.53}$ & $55.30_{\pm1.22}$ & $54.82_{\pm1.02}$ & $64.31_{\pm0.22}$ & $63.80_{\pm0.71}$ & $\textbf{66.30}_{\pm0.51}$ & $65.76_{\pm0.95}$ \\
 RE$_{(9)}$ & $52.27_{\pm2.39}$ & $48.03_{\pm{10.8}}$ & $62.40_{\pm1.41}$ & $60.97_{\pm0.01}$ & $41.01_{\pm0.14}$ & $63.73_{\pm0.70}$ & $65.16_{\pm1.07}$ & $\textbf{65.79}_{\pm0.37}$ & $65.43_{\pm0.93}$ \\
 QA$_{(8)}$ & $50.79_{\pm3.01}$ & $59.56_{\pm0.65}$ & $64.77_{\pm0.65}$ & $59.82_{\pm1.51}$ & $60.98_{\pm0.60}$ & $66.47_{\pm0.56}$ & $\textbf{67.70}_{\pm0.93}$ & $66.82_{\pm0.11}$ & $67.63_{\pm0.86}$ \\
 BERT$_{(10)}$ & $64.95_{\pm0.98}$ & $66.50_{\pm0.09}$ & $66.40_{\pm1.66}$ & $67.15_{\pm0.49}$ & $43.94_{\pm0.70}$ & $68.19_{\pm0.80}$ & $69.06_{\pm0.49}$ & $\textbf{69.07}_{\pm1.15}$ & $68.79_{\pm0.99}$ \\
 Coref$_{(12)}$ & $\textbf{73.75}_{\pm0.29}$ & $72.33_{\pm0.12}$ & $73.75_{\pm0.29}$ & $73.60_{\pm0.31}$ & $73.11_{\pm0.55}$ & $73.60_{\pm0.29}$ & $73.19_{\pm0.65}$ & $72.70_{\pm0.85}$ & $72.59_{\pm0.09}$ \\
\hline
\end{tabular}
}
\caption{Results for single models with different settings, mean and attention aggregation. Subscript indicates the best truncation layer for the \textit{trunc} setting.}\label{tab:single-models}
\end{table*}

%% file: tables/with_variance/2x_models.tex
\begin{table*}[t]
\centering
\scalebox{0.8}{
\begin{tabular}{|l|l|l|l|l|l|l|}
\hline
\multirow{2}*{models} & \multicolumn{3}{|c|}{mean} & \multicolumn{3}{|c|}{attention} \\
\cline{2-7}
 & full & norm & trunc & full & norm & trunc  \\
 \hline
 RE$_{(9)}$ + MRPC$_{(9)}$ & ${61.76_{\pm1.85}}$ & ${64.71_{\pm0.42}}$ & ${64.49_{\pm0.76}}$ & ${67.56_{\pm0.36}}$ & ${51.06_{\pm{13.8}}}$ & ${\textbf{68.78}_{\pm0.32}}$ \\
 NER$_{(6)}$ + MRPC$_{(9)}$ & ${54.71_{\pm4.57}}$ & ${64.47_{\pm0.29}}$ & ${65.71_{\pm0.22}}$ & ${67.94_{\pm0.78}}$ & ${67.36_{\pm0.46}}$ & ${\textbf{68.67}_{\pm0.46}}$ \\
 RE$_{(9)}$ + NER$_{(6)}$ & ${56.13_{\pm1.42}}$ & ${60.68_{\pm0.48}}$ & ${64.38_{\pm1.14}}$ & ${64.00_{\pm0.74}}$ & ${62.80_{\pm0.36}}$ & ${\textbf{67.03}_{\pm0.31}}$ \\
 QA$_{(8)}$ + MRPC$_{(9)}$ & ${60.84_{\pm1.04}}$ & ${65.58_{\pm0.48}}$ & ${66.46_{\pm2.05}}$ & ${67.87_{\pm0.92}}$ & ${59.00_{\pm{15.0}}}$ & ${\textbf{68.98}_{\pm0.44}}$ \\
 RE$_{(9)}$ + QA$_{(8)}$ & ${57.66_{\pm3.61}}$ & ${63.55_{\pm0.26}}$ & ${65.83_{\pm0.61}}$ & ${65.02_{\pm0.39}}$ & ${64.49_{\pm0.30}}$ & ${\textbf{67.86}_{\pm0.59}}$ \\
 NER$_{(6)}$ + QA$_{(8)}$ & ${55.66_{\pm1.72}}$ & ${61.14_{\pm0.74}}$ & ${65.06_{\pm0.68}}$ & ${65.02_{\pm0.46}}$ & ${62.51_{\pm0.59}}$ & ${\textbf{67.65}_{\pm0.55}}$ \\
\hline
\end{tabular}
}
\caption{Results for the pairs of models with different settings, mean and attention aggregation. Subscript indicates the best truncation layer for the \textit{trunc} setting.}\label{tab:2x-models}
\end{table*}

%% file: tables/with_variance/multiple_models.tex
\begin{table*}[t]
\centering
\scalebox{0.8}{
\begin{tabular}{|l|l|l|l|l|l|l|}
\hline
\multirow{2}*{models} & \multicolumn{3}{|c|}{mean} & \multicolumn{3}{|c|}{attention} \\
\cline{2-7}
 & full & norm & trunc & full & norm & trunc  \\
 \hline
 RE$_{(9)}$ + NER$_{(6)}$ + QA$_{(8)}$ & ${58.95_{\pm1.11}}$ & ${63.70_{\pm0.08}}$ & ${65.04_{\pm0.71}}$ & ${66.24_{\pm0.14}}$ & ${64.71_{\pm0.50}}$ & ${\textbf{68.98}_{\pm0.32}}$ \\
 MRPC$_{(9)}$ + NER$_{(6)}$ + QA$_{(8)}$ & ${60.35_{\pm1.42}}$ & ${65.13_{\pm0.65}}$ & ${65.68_{\pm0.83}}$ & ${69.21_{\pm0.17}}$ & ${59.10_{\pm{14.1}}}$ & ${\textbf{69.56}_{\pm0.35}}$ \\
 MRPC$_{(9)}$ + RE$_{(9)}$ + QA$_{(8)}$ & ${61.83_{\pm0.38}}$ & ${65.22_{\pm0.20}}$ & ${66.69_{\pm0.64}}$ & ${68.63_{\pm0.60}}$ & ${67.17_{\pm0.21}}$ & ${\textbf{68.81}_{\pm0.69}}$ \\
 MRPC$_{(9)}$ + RE$_{(9)}$ + NER$_{(6)}$ & ${62.27_{\pm2.08}}$ & ${65.15_{\pm0.18}}$ & ${65.96_{\pm0.52}}$ & ${68.31_{\pm0.10}}$ & ${66.88_{\pm0.16}}$ & ${\textbf{69.30}_{\pm0.52}}$ \\
 MRPC$_{(9)}$ + RE$_{(9)}$ + NER$_{(6)}$ + QA$_{(8)}$ & ${62.19_{\pm1.49}}$ & ${65.56_{\pm0.11}}$ & ${65.66_{\pm0.50}}$ & ${69.03_{\pm0.45}}$ & ${66.80_{\pm0.53}}$ & ${\textbf{69.39}_{\pm0.74}}$ \\
\hline
\end{tabular}
}
\caption{Results for multiple models with different settings, mean and attention aggregation. Subscript indicates the best truncation layer for the \textit{trunc} setting.}\label{tab:multiple-models}
\end{table*}